\documentclass[10pt,twocolumn,letterpaper]{article}

\usepackage{cvpr}              

\usepackage{indentfirst}
\usepackage{capt-of}
\usepackage[dvipsnames]{xcolor}
\usepackage{xspace}
\usepackage{enumitem}
\usepackage{amsmath,amssymb,amsbsy,amsfonts,dsfont,pifont,bm,bbm,mathrsfs,mathtools,nicefrac}
\usepackage{algorithm,algpseudocode,listings}
\usepackage{booktabs,multirow,adjustbox,diagbox,threeparttable,tabularray}

\newcommand{\method}{{GenDeF}\xspace}

\newcommand{\supp}{\textit{Supplementary Material}\xspace}

\newcommand{\tocite}[1]{{\color{red} [TO CITE]}}

\definecolor{linkcolor}{RGB}{255,0,0}
\definecolor{urlcolor}{RGB}{255,105,180}
\definecolor{citecolor}{RGB}{66,168,235}
\definecolor{codegreen}{rgb}{0,0.5,0}
\definecolor{codeblue}{rgb}{0.25,0.5,0.5}
\definecolor{codegray}{rgb}{0.6,0.6,0.6}

\definecolor{cvprblue}{rgb}{0.21,0.49,0.74}
\usepackage[pagebackref,breaklinks,colorlinks,citecolor=cvprblue]{hyperref}
\usepackage{wrapfig}
\usepackage[capitalize]{cleveref}  
\crefname{section}{Sec.}{Secs.}
\Crefname{section}{Section}{Sections}
\crefname{table}{Table}{Tables.}
\Crefname{table}{Table}{Tables}
\crefname{figure}{Fig.}{Figs.}
\Crefname{figure}{Figure}{Figures}
\crefname{equation}{Eq.}{Eqs.}
\Crefname{equation}{Equation}{Equations}

\global\hyphenpenalty=0

\newcommand\nonumfootnote[1]{%
\begingroup%
    \renewcommand\thefootnote{}\footnote{\hspace{-3.7pt}#1}%
    \addtocounter{footnote}{-1}%
\endgroup%
}

\begin{document}

\title{GenDeF: Learning Generative Deformation Field for Video Generation}

\author{
    Wen Wang\textsuperscript{1,2*}, ~~
    Kecheng Zheng\textsuperscript{2}, ~~
    Qiuyu Wang\textsuperscript{2}, ~~
    Hao Chen\textsuperscript{1$\dagger$},   ~~ 
    Zifan Shi\textsuperscript{3,2*},  ~~
    Ceyuan Yang\textsuperscript{4}, ~~ \\[2pt] 
    Yujun Shen\textsuperscript{2$\dagger$}, ~~
    Chunhua Shen\textsuperscript{1}
    \\[6pt]
    $^1$Zhejiang University ~
    $^2$Ant Group ~
    $^3$HKUST
    ~
    $^4$Shanghai Artificial Intelligence Laboratory \\[8pt]
}

\maketitle

\begin{abstract}

We offer a new perspective on approaching the task of video generation.
Instead of directly synthesizing a sequence of frames, we propose to render a video by warping one static image with a generative deformation field (\method).
%
%
Such a pipeline enjoys three appealing advantages.
First, we can sufficiently reuse a well-trained image generator to synthesize the static image (also called canonical image), alleviating the difficulty in producing a video and thereby resulting in better visual quality.
Second, we can easily convert a deformation field to optical flows, making it possible to apply explicit structural regularizations for motion modeling, leading to temporally consistent results.
Third, the disentanglement between content and motion allows users to process a synthesized video through processing its corresponding static image without any tuning, facilitating many applications like video editing, keypoint tracking, and video segmentation.
Both qualitative and quantitative results on three common video generation benchmarks demonstrate the superiority of our \method method. 

Project Page: https://aim-uofa.github.io/GenDeF/

\iftrue
\nonumfootnote{$^*$ Intern at Ant Group.
~~
$^\dagger$ Corresponding authors.
}
\fi

\end{abstract}

\section{Introduction}\label{sec:intro}

Video generation~\cite{saito2017tgan,tulyakov2018mocogan,tian2021mocoganhd,yu2022dign,skorokhodov2022stylegan,zhang2022towards} aims to sample new video from noise, which has a wide range of applications in film production, game development, virtual reality, \textit{etc.} To achieve satisfactory visual effects, the generated videos need to satisfy (1) high quality: each frame in the video 
should be
visually pleasing; (2) temporal consistency: the video should describe the same visual content with reasonable motion patterns.

\begin{figure}
    \centering
    \includegraphics[width=0.48\textwidth]{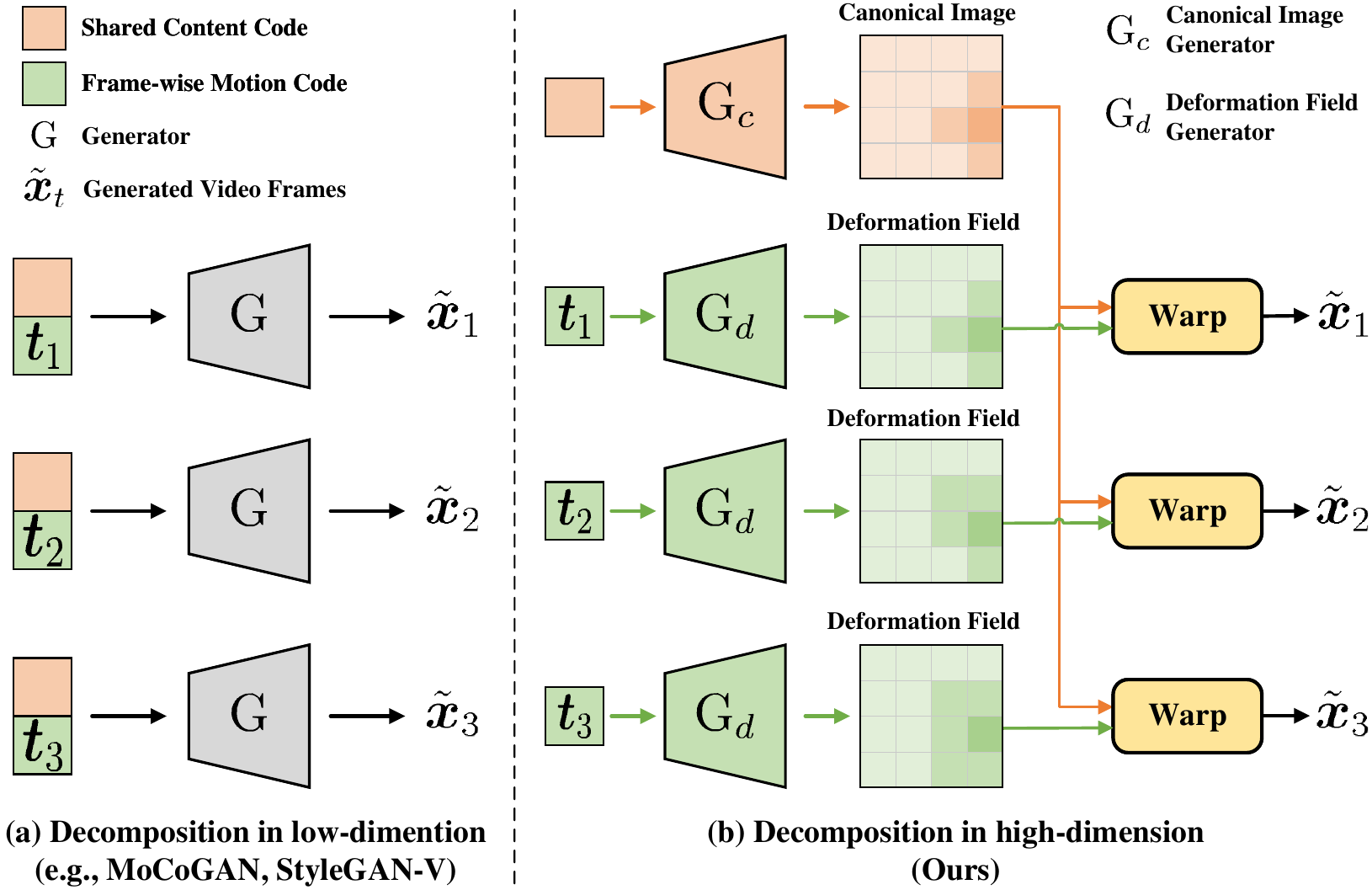}
    \captionsetup{type=figure}
    \caption{%
        \textbf{Paradigm comparison.} (a) Previous methods explore content and motion decomposition in low-dimensional space. (b) We decompose a video into a content-related canonical image and motion-related deformation fields in high-dimensional space with spatial structures, and render the video by warping the canonical image with the deformation fields.
    }
    \label{fig:teaser}
\end{figure}

However, it is not easy to achieve the above goals. As high-dimensional signal 
changes continuously over time, videos contain complex motions, such as camera viewpoint transformation and deformation of objects. Considering the spatial redundancy of repeated content in videos, existing video generation methods \cite{tulyakov2018mocogan,skorokhodov2022stylegan} 
mitigate 
the difficulty of learning video generation by decomposing the latent space of videos into low-dimensional content latent variables shared by different frames and motion latent variables that vary across time, as shown in Fig.~\ref{fig:teaser}a. Despite the progress, the content and motion in these approaches are decoupled only in a low-dimensional latent space, making it difficult for interpretation or downstream applications.

In this paper, we propose a novel video generation paradigm, termed \textbf{\method}, which explicitly decomposes the video into a content-related canonical image and a motion-related deformation field in a high-dimensional space with spatial structures, as shown in \cref{fig:teaser}b. The video can be obtained by warping the shared canonical image with frame-wise deformation fields.
Such an explicit decomposition allows us to fully reuse the pixel information in the canonical image, significantly mitigating the difficulty of independently predicting the pixel values in each video frame, which leads to improved generation quality.
Moreover, unlike low-dimensional latent variables, our canonical image and deformation field are spatially structured, making it possible to utilize explicit structural regularization to improve video generation performance. In practice, we can easily obtain optical flows from the deformation field and apply explicit structural regularization on optical flows to improve the temporal consistency of the generated video.

Experiments on several commonly used video generation benchmarks show that \method achieves state-of-the-art video generation results,
surpassing previous methods in both temporal consistency and the visual quality of individual frames.
Notably, thanks to the decomposition of content and deformation field, we can generate multiple plausible videos of the same visual content performing varying motions.
Moreover, our method facilitates various downstream video processing applications. Specifically, users can directly perform image processing on the interaction-friendly canonical image. Subsequently, the deformation field can smoothly propagate the image processing on the canonical image to the whole video, realizing consistent video editing, point tracking, video segmentation, 
and more.
\section{Related Work}\label{sec:related}

\noindent\textbf{Video synthesis.}
Many works explore GAN-based video generation. Early work such as TGAN~\cite{saito2017tgan} and VGAN~\cite{vondrick2016generating} extend 2D GANs~\cite{gan,karras2017progressive,karras2019stylegan,karras2020stylegan2,karras2021alias} to video by assuming that each video clip corresponds to a point in a video latent space. MoCoGAN~\cite{tulyakov2018mocogan} argued that such an assumption dramatically improves the difficulty of video learning, and assumes that a video clip is obtained by traversing the points in an image latent space. To this end, MoCoGAN decouples content and motion in the latent space. MoCoGAN-HD~\cite{tian2021mocoganhd} and StyleInv~\cite{wang2023styleinv} follow the idea and use autoregressive and non-autoregressive approaches, respectively, to find motion trajectories in the latent space to generate videos. MoStGAN-V~\cite{shen2023mostgan} further improves motion modeling by introducing a motion style attention modulation. 
Considering the intrinsic continuous property of video, some recent works~\cite{yu2022dign,skorokhodov2022stylegan} introduce neural implicit representations into video generation. For example, DIGAN~\cite{yu2022dign} represents a video as a set of MLPs and samples the spatial-temporal coordinates to produce videos in pixel space. Similarly, StyleGAN-V~\cite{skorokhodov2022stylegan} samples motion codes from a time-continuous latent space. Further, StyleSV~\cite{zhang2022towards} and LongVideoGAN~\cite{brooks2022generating} explore long video generation, and StyleSV-MTM~\cite{yang2023learning} introduces a modulated transformation module to handle video geometry deformations.

Several works explore video generation based on other generative models. For example, VideoGPT~\cite{yan2021videogpt} and TATS~\cite{ge2022long} use VQ-VAE~\cite{van2017neural} or VQ-GAN~\cite{esser2021taming} to compress the video into a discrete latent space, and then leverage Transformer~\cite{vaswani2017attention} for autoregressive video generation. Recently, some works~\cite{ho2022videodiffusion,yang2022diffusion,ho2022imagen,singer2022make,zhou2022magicvideo,blattmann2023align,ge2023preserve,wang2023videocomposer} explore video generation based on diffusion models, \textit{e.g}., VDM~\cite{ho2022videodiffusion} extends image diffusion models to video generation by introducing joint image-video training and a 3D UNet structure. To improve the efficiency, PVDM~\cite{yu2023video} compresses the video to factorized 2D latents to ease the optimization. Compared to GANs, these methods typically require larger training costs and a longer sampling time to obtain visually pleasing results.

We build our method based on GANs. Unlike existing methods~\cite{tulyakov2018mocogan,skorokhodov2022stylegan} that decouple content and motion in the latent space, we explicitly decompose the video into pixel-space visual contents and structured deformation fields. Such decomposition drastically reduces the difficulty of learning content and motion simultaneously, makes explicit structural regularizations possible, and facilitates applications like video editing, point tracking, \textit{etc}.

\noindent\textbf{Image-to-video generation.}
The goal of image-to-video generation~\cite{blattmann2021ipoke,blattmann2021understanding,dorkenwald2021stochastic,hu2022make,mahapatra2022controllable,pan2019video,wang2020imaginator,xiong2018learning,yang2018pose,zhang2020dtvnet,ni2023conditional} is to generate plausible videos from an input image. Existing methods can be roughly divided into two categories. While one line of works generates raw pixels from the observed input image~\cite{wang2020imaginator,xiong2018learning,dorkenwald2021stochastic,yang2018pose}, the other line predicts spatial transformations that warp the input image to future frames~\cite{li2018flow,pan2019video,zhang2020dtvnet}. 
Although the latter works share some similarities with our \method in that both obtain video frames by warping an image, there are several fundamental differences.
First, image-to-video generation is a conditional generation task, whereas we tackle unconditional generation, which generates videos from scratch.
Second, the canonical image in our \method is not a natural image, but a compressed presentation of the content information in the entire video. 
Thirdly, our experiments in \cref{subsec:exp_ablate} also show that replacing the canonical image with images generated by a pre-trained image generator results in poor video generation.

\noindent\textbf{Neural atlases for video decomposition.}
Our work is inspired by the recent study of neural atlases~\cite{LNA, deformable_sprites, codef, inve, lu2020layered, lu2022associating, rav2008unwrap, shade1998layered}, which involves the decomposition of videos into a canonical image through the learned deformation fields.
Layered Neural Atlas~\cite{LNA} implements an implicit network, specifically designed to discern between foreground and background movement, effectively partitioning them into separate layers.
To better capture non-rigid motions within the video content, Deformable Sprites~\cite{deformable_sprites} conceptualizes the transformation as a composition of a homography intertwined with 2D spatial splines that evolve smoothly over time.
However, these approaches take too much time for optimization on individual videos.
CoDeF~\cite{codef} and INVE~\cite{inve} enhance both model capacity and speed by integrating hash-grids encoding~\cite{muller2022instant}.

While the aforementioned methodologies employ an innovative video representation to achieve consistent video editing, the primary focus of our work leans toward video generation.
We integrate deformation fields into our generation procedure, deviating from the conventional approach of decoding individual video frames independently.
As demonstrated below, this design offers an alternative and superior approach to video generation.

\begin{figure*}
    \centering
    \includegraphics[width=\textwidth]{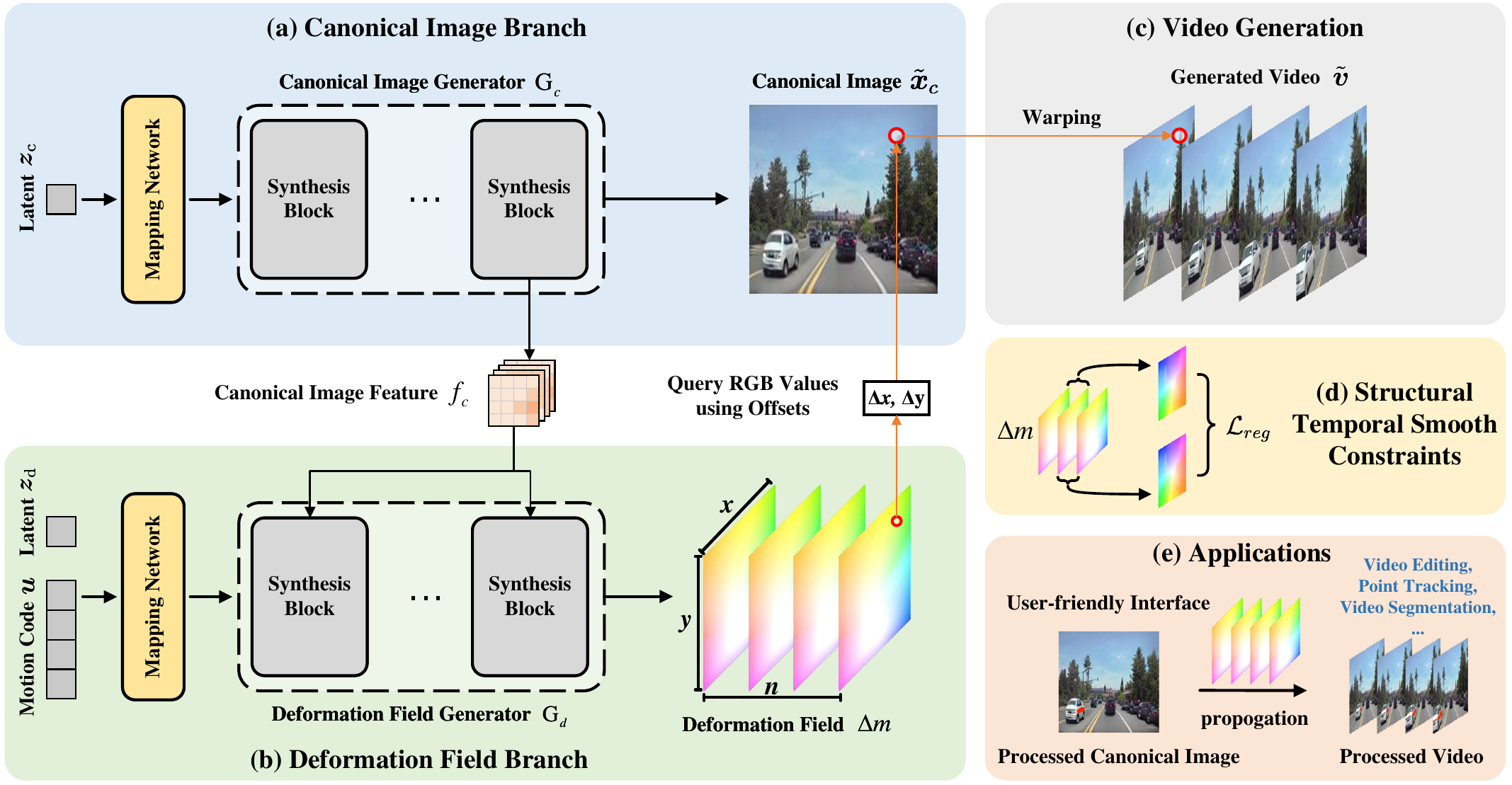}
    \captionsetup{type=figure}
    \caption{%
        \textbf{Illustration of our \method}. We decompose the video into a content-related canonical image in (a) and a frame-wise deformation field in (b). The video can be obtained by warping the canonical image using the deformation field (c). The explicit decomposition enables structural temporal smoothness regularization in (d) and a variety of downstream applications in (e).
        Both the canonical image and deformation field generators follow the StyleGAN~\cite{karras2019stylegan} structure, with a mapping network and multiple synthesis blocks.
    }
    \label{fig:method}
\end{figure*}

\section{Method}\label{sec:method}

Given a video dataset, our goal is to learn a generative model over the data distribution. 
To this end, we propose a novel video generation paradigm that explicitly decouples the video into a canonical image shared by all frames and the deformation field that warps the canonical pixels to each video frame. 
The overview of our \method is shown in~\cref{fig:method}. 
We elaborate the model design in~\cref{subsec:method_model}, and the training objectives in~\cref{subsec:method_loss}. Afterward, we show how \method facilitates various video processing applications in~\cref{subsec:method_apply}.

\subsection{\method Model Design}
\label{subsec:method_model}

Our \method is a GAN-based video generation framework, which consists of a canonical image branch $\mathrm{G}_c$ for canonical image generation, a deformation field branch $\mathrm{G}_d$ for deformation field generation, and a discriminator $\mathrm{D}$ for discriminating real videos from the dataset and the generated fake videos.

\noindent\textbf{Canonical image branch.}
Inspired by neural atlases~\cite{codef, inve, shade1998layered}, we also utilize the canonical image to represent the entire video content. 
Specifically, we first sample a canonical latent code $\boldsymbol{z}_{\mathrm{c}}$ from the noise distribution, \textit{i.e.}, $\boldsymbol{z}_{\mathrm{c}} \sim \mathcal{N}(\boldsymbol{0}, \boldsymbol{I})$. 
Subsequently, the canonical image generator $\mathrm{G}_c$ takes $\boldsymbol{z}_{\mathrm{c}}$ as input and generates a canonical image. 
The process can be written as:
\begin{equation}
\tilde{\boldsymbol{x}}_c=\mathrm{G}_c\left(\boldsymbol{z}_c\right),
\end{equation}
where $\tilde{\boldsymbol{x}}_c \in \mathbb{R}^{H \times W \times 3}$ is the canonical image. $H$ and $W$ represent the height and width of the canonical image, respectively.
The canonical image should contain the content information of the entire video in pixel space and hence it is time-independent. Since a large amount of redundancy exists among different frames in a video, the content of an individual video frame can be a good initialization for the canonical image. Therefore, we pre-train the canonical image generator with individual video frames from the dataset.

\noindent\textbf{Deformation field branch.}
The deformation field encodes the correspondence between pixels in the canonical image and those in video frames at individual times, and thus it should be time-dependent. Inspired by MoCoGAN~\cite{tulyakov2018mocogan}, we decompose the latent space of the deformation field into a global $\boldsymbol{z}_{\mathrm{c}} \sim \mathcal{N}(\boldsymbol{0}, \boldsymbol{I})$ and a time-dependent motion code $\boldsymbol{u}_{\mathrm{t}}$. We follow StyleGAN-V~\cite{skorokhodov2022stylegan} to obtain $\boldsymbol{u}_{\mathrm{t}}$, where we encode the equidistant sampled noise with a lightweight motion mapping network and then apply acyclic positional encodings.

The deformation field generated by $\mathrm{G}_d$ should reflect the motion pattern of objects in the canonical image. In other words, the deformation field corresponding to a stationary mountain in the canonical should be different from that corresponding to a moving person. To maintain the consistency between the deformation field and the corresponding canonical image, we insert the intermediate features from the canonical image generator into the deformation field generator, as shown in~\cref{fig:method}. In this way, the generated deformation fields can be canonical-aware. The generation process of the deformation field can be written as:
\begin{equation}
\Delta\boldsymbol{m}_t=\mathrm{G}_d\left(\boldsymbol{z}_{d}, \boldsymbol{u}_{t} ; \boldsymbol{f}_{c}\right), t=1,2, \ldots, n,
\end{equation}
where $\boldsymbol{f}_{c}$ is the inserted canonical feature, $n$ is the number of frames in the video, and $\Delta\boldsymbol{m}_t \in \mathbb{R}^{H \times W \times 2}$ is the deformation filed at the $t$-th frame.

Concretely, we interpolate $\boldsymbol{f}_{c}$ into the same resolution as the feature from the deformation field generator, before concatenating them at the beginning of each synthesis block of $\mathrm{G}_d$. 
We use the feature of the penultimate layer in $\mathrm{G}_c$ as the canonical feature, since the near-image output feature has a clearer structure compared to the early layers and enjoys less information compression compared to the output layer. 

To ease the deformation field learning, we use the spatial coordinates of pixels in the output image as references, denoted as $\boldsymbol{m}$. Instead of directly learning the coordinates of the pixels in the canonical image, our deformation field stores the coordinates offsets between the canonical pixels and the corresponding output pixels. After generating the deformation field, we can obtain the video frames by
\begin{equation}
\tilde{\boldsymbol{x}}_{t}=\mathcal{T}\left(\tilde{\boldsymbol{x}}_c, \boldsymbol{m} + \Delta\boldsymbol{m}_t\right), t=1,2, \ldots, n,
\end{equation}
where $\mathcal{T}$ is the warping function that maps each pixel in the canonical image to the corresponding locations in output images, $\tilde{\boldsymbol{x}}_{t}$ is the $t$-th generated video frame, and the generated video can be denoted as $\tilde{\boldsymbol{v}}=\left\{\tilde{\boldsymbol{x}}_1, \tilde{\boldsymbol{x}}_2, \cdots, \tilde{\boldsymbol{x}}_n\right\}$.

\subsection{Training Losses}
\label{subsec:method_loss}

\noindent\textbf{Structural temporal smoothness constraint.}
In \method, the canonical image encodes time-independent content information, while the deformation field explicitly encodes time-dependent motion in the video. Unlike previous approaches~\cite{tulyakov2018mocogan,skorokhodov2022stylegan} that decompose content and motion implicitly in the latent space, our decomposed deformation field contains explicit spatial structures, which allows us to apply structural constraints on the deformation field to improve the temporal consistency of the generated video.
Ideally, we would like to generate videos where the temporal differences between neighboring frames are continuous. To achieve this goal, we assume that the optical flows between neighboring frames are similar. The optical flow between neighboring frames can be estimated easily from the deformation field, \textit{i.e.}, $\mathcal{F}_{t \rightarrow t+1}=m_{t+1}-m_{t}$. Based on the estimated optical flow, our structural temporal smoothness constraint can be expressed as
\begin{equation}
\mathcal{L}_{reg} = \operatorname{Sim}(\mathcal{F}_{t \rightarrow t+1}, \mathcal{F}_{t+1 \rightarrow t+2}),
\end{equation}
where $\operatorname{Sim}$ computes the similarity between optical flows. Although more sophisticated metrics can be applied, we empirically find that the simple edge-aware smoothness loss~\cite{godard2017unsupervised} works sufficiently well.

\noindent\textbf{Overall objectives.}
To generate videos that comply with the distribution of the dataset, we train \method with the adversarial losses in StyleGAN-V~\cite{skorokhodov2022stylegan}, \textit{i.e.},
\begin{equation}
    \mathcal{L}_{adv} = \mathbb{E}_{\boldsymbol{v}}[-\log \mathrm{D}(\boldsymbol{v})]+\mathbb{E}_{\tilde{\boldsymbol{v}}}[-\log (1-\mathrm{D}(\tilde{\boldsymbol{v}}))],
\end{equation}
where $\tilde{\boldsymbol{v}}$ and $\boldsymbol{v}$ are generated videos and real videos in the dataset, respectively.
The overall training objective for training our generators and the discriminator is
\begin{equation}
\max _{\mathrm{G}_{c}, \mathrm{G}_d} \min _\mathrm{D}  \mathcal{L}_{adv} + \min _{\mathrm{G}_{c}, \mathrm{G}_d}\lambda_{reg} \mathcal{L}_{reg}.
\end{equation}
Here $\lambda_{reg}$ is the hyperparameter that balances the two losses.

\subsection{Propagating Canonical to Video}
\label{subsec:method_apply}

During the sampling process, our \method can not only generate the video, but also produce both the canonical image containing the content of the video and the deformation field containing the per-pixel correspondence between each video frame and the canonical image. What's more, benefiting from the image pre-training initialization, the obtained canonical image is close to a natural image, making it easy to both interpret and manipulate. 

These features allow us to manipulate the canonical image and subsequently propagate it throughout the video, thus facilitating the following applications. (1) Consistent video editing: we can propagate the manual or algorithm-generated editings on the canonical image to the whole video. (2) Point tracking: we can select the coordinates of a point of interest in the canonical image and use the deformation field to obtain the trajectory of the point in the entire video. (3) Video segmentation: we can generate the segmentation mask of the object of interest on the canonical image, and then propagate it to obtain the video segmentation result.

Moreover, thanks to our decomposition, we can vary the latent code $\boldsymbol{z}_d$ and motion code $\boldsymbol{u}$ to generate multiple plausible deformation fields for the same canonical image, thus obtaining multiple videos of the same visual content performing different motions.
\section{Experiments}\label{sec:exp}

\begin{figure*}
    \centering
    \includegraphics[width=\textwidth]{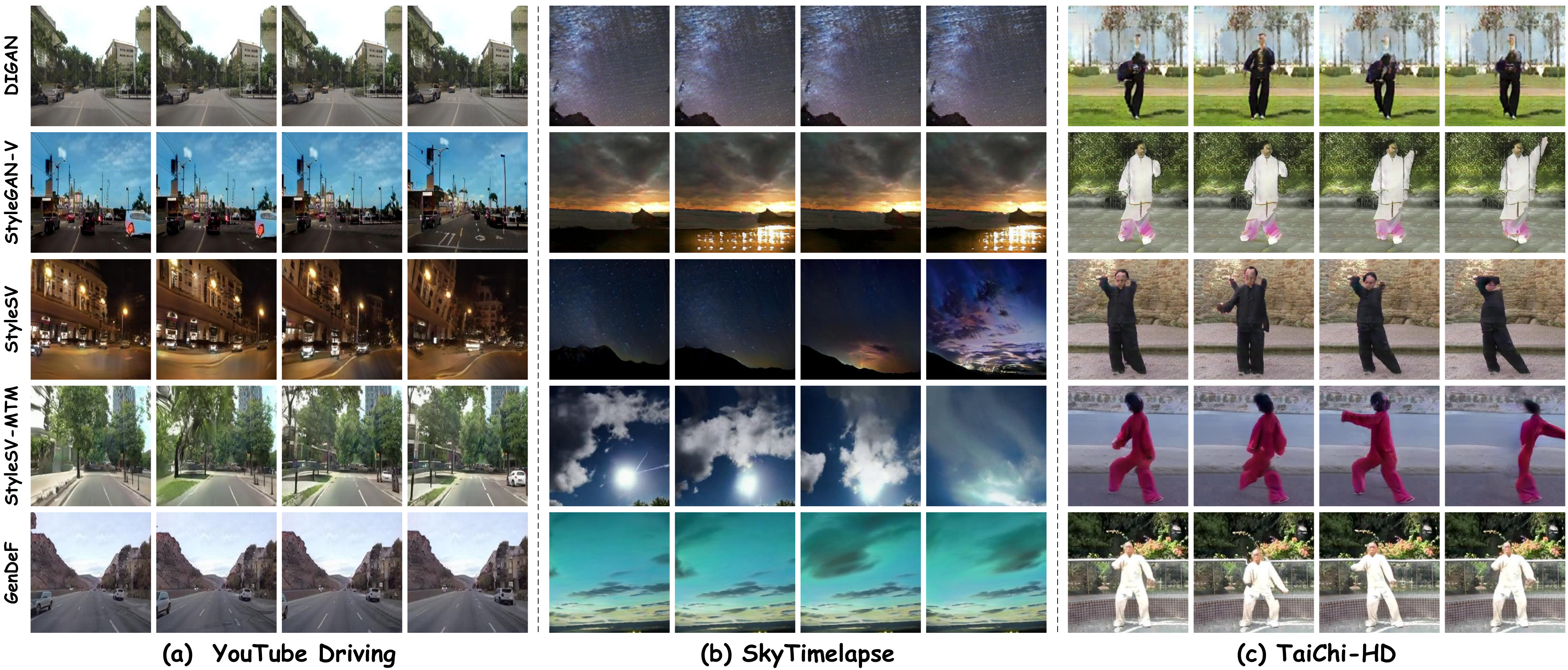}
    \caption{%
        \textbf{Qualitative comparison.} The video samples are taken from (a) YouTube Driving, (b) SkyTimelapse, and (c) TaiChi-HD, respectively. Due to space constraints, we only show 4 evenly-spaced frames of the generated 128-frame videos.
    }
    \label{fig:main}
\end{figure*}

\subsection{Experimental Setups}
\label{subsec:exp_setups}

\noindent\textbf{Datasets.}
We conduct experiments on three commonly used video datasets to validate the effectiveness of our method, including (1) SkyTimelapse~\cite{xiong2018learning}, which contains scenes of clouds, cloud shadows, aurora, \textit{etc.}, captured by a stationary camera. (2) TaiChi-HD~\cite{Siarohin_2019_NeurIPS}, which contains videos of people playing taichi in various scenes. (3) YouTube Driving~\cite{zhang2022learning}, which contains first-perspective videos of driving scenes with diverse weather, scenes, and cities. We follow existing works~\cite{skorokhodov2022stylegan,zhang2022towards} to resize the videos to 256$\times$256 resolution.

\noindent\textbf{Evaluation metrics.}
Following the literature~\cite{skorokhodov2022stylegan,zhang2022towards}, we use Frechet Video Distance (FVD)~\cite{unterthiner2018towards} and Frechet Image Distance (FID)~\cite{heusel2017gans} for quantitative verification. Specifically, we generate 2,048 videos of 128 frames and use all 128 video frames or the first 16 frames to compute FVD$_{128}$ and FVD$_{16}$, respectively. In addition, we sample 50k video frames from the real and generated videos to compute the FID, which measures the quality of single frames in a video.

\noindent\textbf{Implementation details.}
All experiments are conducted on 8 A100 GPUs. We first pre-train the canonical image generator and the discriminator with individual frames from the video dataset, followed by the joint fine-tuning of the entire model on videos. In the fine-tuning stage, we initialize the deformation field to zeros. We follow StyleGAN-V's sparse training strategy, which randomly samples only three frames from the video at each training iteration. $\mathrm{G}_c$ is implemented using StyleGAN-3~\cite{karras2021alias} structure, while $\mathrm{G}_d$ and $\mathrm{D}$ follow the structure of StyleGAN-V~\cite{skorokhodov2022stylegan}. We adopt the modulated transformation module~\cite{yang2023learning} in both generators. $\lambda_{reg}$ is set to 1 across datasets. Please refer to \supp for more details.

\subsection{Main Results}
\label{subsec:exp_main}

\noindent\textbf{Quantitative results.}
\Cref{table:main} shows our quantitative results. It can be seen that our method achieves the best results on FVD$_{16}$ and FID across all three datasets. For FVD$_{128}$, our method achieves the best results on TaiChi-HD and YouTube Driving, and very competitive results on SkyTimelapse. These results indicate that our method can achieve better results in terms of overall video quality and temporal consistency. It is worth mentioning that our method significantly improves the quality of single image frames in the video. We speculate that the reuse of the pixel content in canonical images reduces the difficulty of learning video frames individually, thus leading to better visual quality.

\noindent\textbf{Qualitative results.}
Fig.~\ref{fig:main} shows the video samples generated by the different methods. We have the following observations: (1) DIGAN and StyleGAN-V often produce unrealistic video results accompanied by severe artifacts, as shown in the sky and lake regions on SkyTimelapse. (2) StyleSV and StyleSV-MTM can obtain slightly better results, but they often have difficulties in maintaining temporal consistency, \textit{e.g.}, the sky color at the beginning and end of videos on SkyTimelapse differs significantly. (3) Our method can produce temporally consistent video samples, while also enjoying much higher single-frame image quality, \textit{e.g.}, our results on TaiChi-HD demonstrate more details in both the foreground person and the background regions. This is consistent with the superior performance we observed on the FID metric.

\begin{table}[t]
    \caption{
    \textbf{Quantitative comparisons on video generation.} FVD$_{16}$ ($\downarrow$), FVD$_{128}$ ($\downarrow$), and FID ($\downarrow$) are reported for the quantitative measurement. The best results are highlighted in \textbf{bold}.
    }
    \label{table:main}
    \centering\footnotesize
    \subfloat[Evaluation on SkyTimelapse~\cite{xiong2018learning}.]
    {
        \begin{tabular}{lccc}
        \toprule
        Methods                                    & FVD$_{16}$ & FVD$_{128}$ & FID    \\
        \midrule
        MoCoGAN~\citep{tulyakov2018mocogan}        &  85.9      & 272.8       &  -     \\
        DIGAN~\citep{yu2022dign}                   &  83.1      & 196.7       &  -     \\
        LongVideoGAN~\citep{brooks2022generating}  & 116.5      & 152.7       &  -     \\
        TATS-base~\citep{ge2022long}                        & 132.6      &  -          &  -     \\
        StyleGAN-V~\citep{skorokhodov2022stylegan} &  73.9      & 248.3       &  40.8  \\
        
        PVDM~\cite{yu2023video}                             &  55.4     & 125.2       &  -     \\
        StyleSV~\cite{zhang2022towards}            &  49.0      & 135.9       &  49.9  \\
        StyleSV-mtm~\cite{yang2023learning}                      &  42.3      & \textbf{124.6 }      &  44.5  \\ 
        \midrule
        \method                                    & \textbf{41.8} & {128.1} & \textbf{40.7}  \\ 
        \bottomrule
        \end{tabular}
    }
    \hfill
  
    \subfloat[Evaluation on TaiChi-HD~\cite{Siarohin_2019_NeurIPS}.]
    {
        \begin{tabular}{lccc}
        \toprule
        Methods                                    & FVD$_{16}$ & FVD$_{128}$ & FID     \\
        \midrule
        MoCoGAN-HD~\citep{tian2021mocoganhd}       & 144.7      &  -          &  -      \\
        DIGAN~\citep{yu2022dign}                   & 128.1      &  -          &  -      \\
        TATS-base~\citep{ge2022long}               &  94.6      &  -          &  -      \\
        StyleGAN-V~\citep{skorokhodov2022stylegan} & 152.0      & 267.3       &  33.8   \\
        StyleSV~\cite{zhang2022towards}            &  97.4      & 188.9       &  26.6   \\
        StyleSV-mtm~\cite{yang2023learning}                      &  89.5      & 180.6       &  25.2   \\ 
        \midrule
        \method                                    & \textbf{88.2}  & \textbf{172.4}   & \textbf{20.7}\\ 
        \bottomrule
        \end{tabular}
    }
  
    \subfloat[Evaluation on YouTube Driving~\cite{zhang2022learning}.]
    {
        \begin{tabular}{lccc}
        \toprule
        Methods      & FVD$_{16}$      & FVD$_{128}$ & FID                             \\
        \midrule
        DIGAN~\citep{yu2022dign}  & 415.1 & 412.7 & 26.6 \\
        StyleGAN-V~\citep{skorokhodov2022stylegan}           &     449.8     &  460.6    & 28.3    \\
        StyleSV~\cite{zhang2022towards} & 207.2 & 221.5 & 19.2 \\
        StyleSV-mtm~\cite{yang2023learning} & 194.8 & 198.4 & 10.3  \\ 
        \midrule
        \method  & \textbf{181.9} & \textbf{188.5} & \textbf{8.0}  \\ 
        \bottomrule
        \end{tabular}
    }
\end{table}

\subsection{Ablation Study}
\label{subsec:exp_ablate}

\begin{figure}
    \centering
    \includegraphics[width=0.48\textwidth]{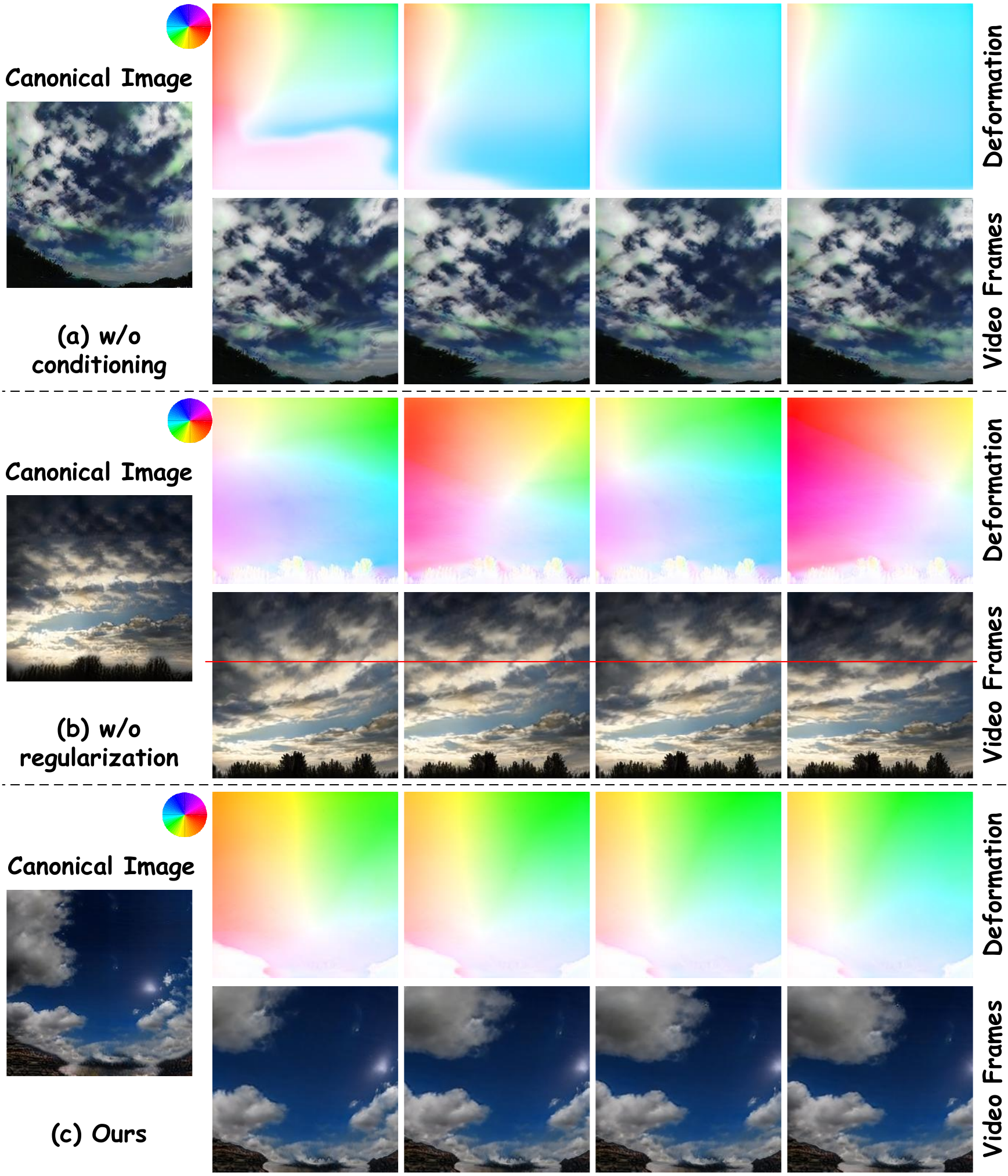}
    \captionsetup{type=figure}
    \caption{%
        \textbf{Ablations on each module.}
        Similar to Table~\ref{table:ablate_module}, we remove the conditional canonical feature in the deformation field generator in (a), and remove the structural temporal smoothness constraint in (b). Our result is shown in (c). 
        Canonical images, deformation fields, and the corresponding video frames are visualized for comparison.
    }
    \label{fig:ablate_module}
\end{figure}

\begin{table}[t]
\footnotesize
\caption{
    \textbf{Ablations on each module.} 
    ``\textit{w/o} $\boldsymbol{f}_{c}$" indicates removing the conditional canonical feature in the deformation field generator, and ``\textit{w/o} $\mathcal{L}_{reg}$" indicates removing the structural temporal smoothness constraint. 
}
\label{table:ablate_module}
\vspace{2pt}
\centering
\begin{tabular}{@{}lllllll@{}}
\toprule
\multirow{2}{*}{\textbf{Config}}              & \multicolumn{3}{c}{YouTube Driving} & \multicolumn{3}{c}{SkyTimelapse} \\ \cmidrule(l){2-7} 
            & FVD$_{16}$    & FVD$_{128}$   & FID           & FVD$_{16}$    & FVD$_{128}$   &  FID       \\ 
\midrule
\textit{w/o} $\boldsymbol{f}_{c}$    & 553.4        & 407.8     & 13.4          &  87.3         & 232.1         & 47.1        \\
\textit{w/o} $\mathcal{L}_{reg}$     & 191.4        & 190.6     & \textbf{7.9}  & \textbf{40.8} & 144.9         & \textbf{40.0} \\ 
\midrule
\method                              & \textbf{181.9} & \textbf{188.5} & {8.0}  & {41.8}        & \textbf{128.1}       & {40.7} \\
\bottomrule
\end{tabular}
\end{table}

\noindent\textbf{Ablations on each module.}
We conduct ablated experiments to validate the effectiveness of the key designs in \method, as shown in \Cref{table:ablate_module} and Fig.~\ref{fig:ablate_module}. 
First, we remove the conditional canonical feature $\boldsymbol{f}_{c}$ in the deformation field generator, denoted as ``\textit{w/o} $\boldsymbol{f}_{c}$", it can be seen that the generation quality drops dramatically without conditioning on the canonical feature. As evidenced by Fig.~\ref{fig:ablate_module}a, the deformation field generated by $\mathrm{G}_d$ is irrelevant to the canonical image, and even the mountains in the background are moving, resulting in an unrealistic video. 
Secondly, we remove the structural temporal smoothness constraint to ablate its effectiveness, denoted as ``\textit{w/o} $\mathcal{L}_{reg}$". Although the quantitative improvement brought by the regularization is limited, we can see from Fig.~\ref{fig:ablate_module}b that abrupt changes occur in the generated deformation field. As a result, discontinuous object motions can be observed in the video, and the video clouds move back and forth in the video. In contrast, the motion of the foreground region is continuous in our result, while the background region is kept stationary, as shown in Fig.~\ref{fig:ablate_module}c.

\begin{table}[t]
\footnotesize
\caption{
    \textbf{Ablations on canonical feature for conditioning.} \#L indicates the layer index in the canonical image generator $\mathrm{G}_{c}$.
}
\label{table:ablate_cond}
\vspace{2pt}
\centering
\begin{tabular}{@{}cccc|cccc@{}}
\toprule
\textbf{\#L}  & FVD$_{16}$   & FVD$_{128}$   & FID    & \textbf{\#L}    & FVD$_{16}$    & FVD$_{128}$   &  FID       \\ 
\midrule
4 & 284.0 & 237.9 & 11.1 & 12 & 189.5 & 202.4 & \textbf{7.6} \\
8 & 265.3 & 227.5 & 10.3 & 13 & \textbf{181.9} & \textbf{188.5} & {8.0} \\
10 & 241.2 & 198.4 & 10.8 & 14 & 363.7 & 382.4 & 12.9 \\
\bottomrule
\end{tabular}
\end{table}

\noindent\textbf{Ablations on canonical conditioning strategy.}
As described in Sec.~\ref{subsec:method_model}, we use the penultimate feature in the canonical image generator as the conditional feature $\boldsymbol{f}_{c}$ for generating deformation fields. Here we ablate this design choice and the results are shown in Table~\ref{table:ablate_cond}. 
It can be seen that from the 4th to the 13th layer, using canonical features from deeper layers consistently improves the overall performance. 
We believe this is mainly because the features closer to the output layer have a higher resolution and contain more explicit structural information, which helps align the structure of the deformation field with that of the canonical image. 
However, directly using the output canonical image from the 14th layer as the conditional feature produces poor results. We speculate that the output canonical image deeply compresses the canonical information to only 3 channels, making it difficult to learn the corresponding deformation field.

\begin{table}[t]
\footnotesize
\caption{
    \textbf{Ablations on the training strategy.} ``\textit{w/o} pt." indicates removing the image-level pre-training for the canonical image generator, and ``pt. then fix" represents freezing the pre-trained canonical image generator during the fine-tuning stage.
}
\label{table:ablate_pt}
\vspace{2pt}
\centering
\begin{tabular}{@{}lllllll@{}}
\toprule
\multirow{2}{*}{\textbf{Config}}              & \multicolumn{3}{c}{YouTube Driving} & \multicolumn{3}{c}{SkyTimelapse} \\ \cmidrule(l){2-7} 
            & FVD$_{16}$    & FVD$_{128}$   & FID           & FVD$_{16}$    & FVD$_{128}$   &  FID       \\ 
\midrule
\textit{w/o} pt.    & 876.4 & 694.6 & 51.4 & 115.7 & 275.4 & 63.6 \\
pt. then fix  & 354.1 & 337.7 & 17.3 & 64.5 & 190.2 & 47.1 \\ 
\midrule
\method     & \textbf{181.9} & \textbf{188.5} & \textbf{8.0}       & \textbf{41.8}        & \textbf{128.1}       & \textbf{40.7} \\
\bottomrule
\end{tabular}
\end{table}

\noindent\textbf{Ablations on the training strategy.}
By default, we perform image-level pre-training of the canonical image generator with individual video frames, before fine-tuning the entire \method model on the video dataset. Here we ablate this design and the results are shown in Table~\ref{table:ablate_pt}. It can be seen that after removing the pre-training on the canonical image generator, the model performance drops dramatically, \textit{e.g.}, from 181.9 to 876.4 on FVD$_{16}$ for the YouTube Driving dataset. We believe this is caused by the difficulty of learning both canonical images and deformation fields from scratch. 
Second, we fix the canonical image generator during the fine-tuning phase. In this way, the model predicts the deformation field to warp a single image to generate the video. It can be seen that the model performance decreases significantly in both image and video qualities, as reflected by the obvious increase in FID and FVD metrics. 
These results indicate that the canonical image can not be replaced by a single generated image.
We suspect a single image contains limited content information and it is difficult to reuse pixels from it to form a complete video. Differently, our method fine-tunes the canonical image generator and allows the model to make adjustments to the canonical image and encode more information to form the video content.

\subsection{Applications}
\label{subsec:exp_apply}

\begin{figure}
    \centering
    \includegraphics[width=0.48\textwidth]{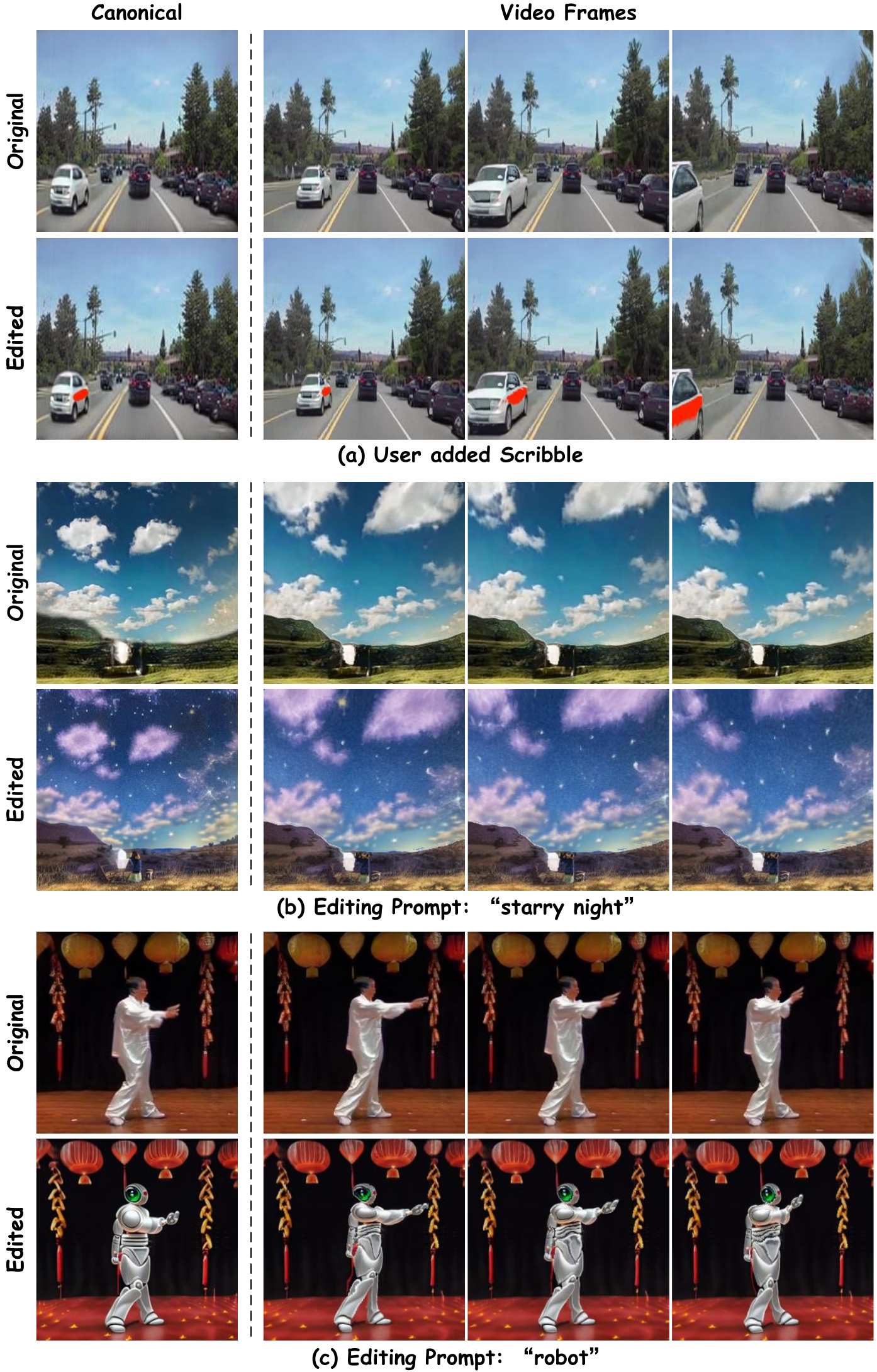}
    \caption{%
        \textbf{Applications on video editing.}
        Note the edits are only applied to the canonical image before being automatically propagated to the video frames.
        We consider (a) user add scribbles, (b) style editing, and (c) object editing.
    }
    \label{fig:edit}
\end{figure}

\begin{figure*}
    \centering
    \includegraphics[width=\textwidth]{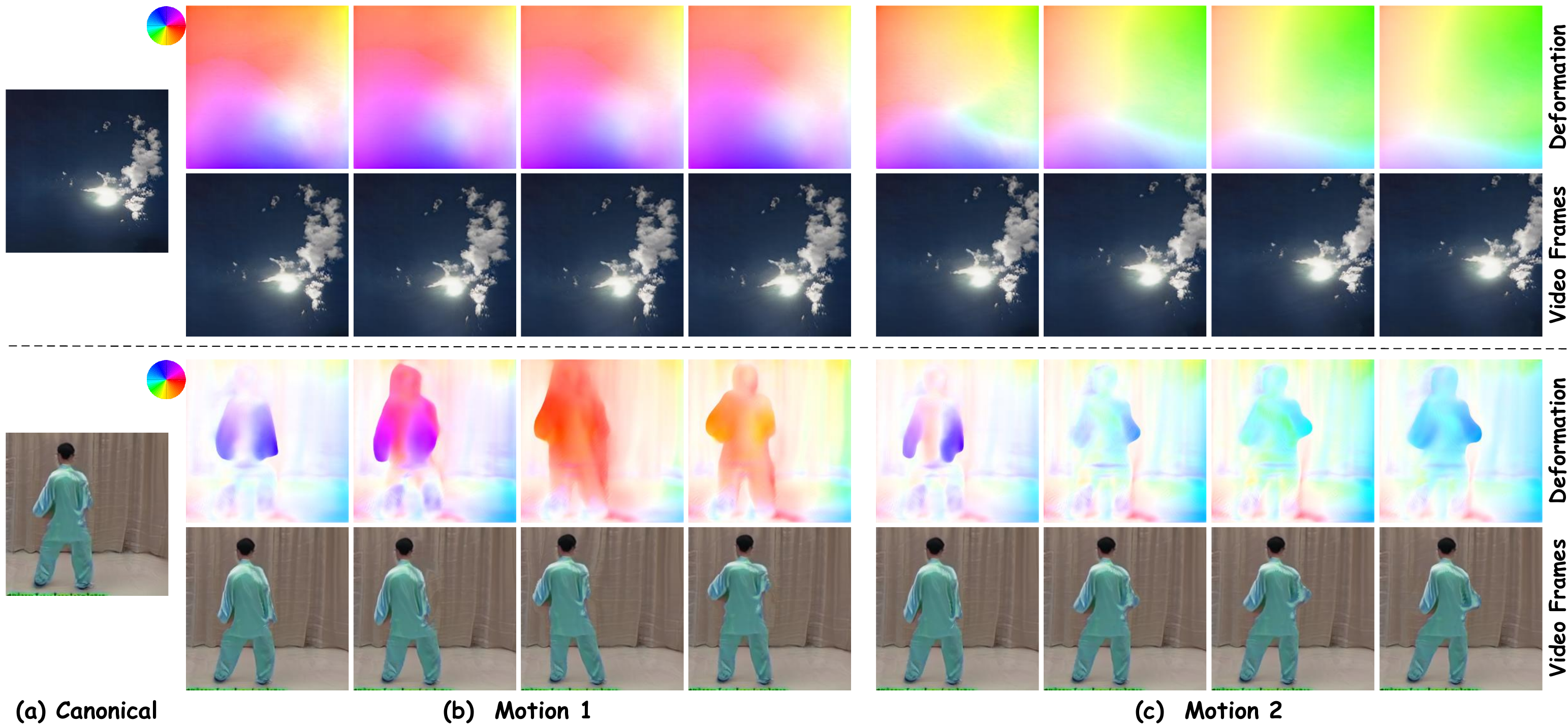}
    \captionsetup{type=figure}
    \caption{%
        \textbf{Sampling multiple plausible motions for the same visual content.} The canonical image is shown in (a), while two samples with different motions are shown in (b) and (c), respectively.
    }
    \label{fig:multi_motion}
\end{figure*}

\begin{figure}
    \centering
    \includegraphics[width=0.48\textwidth]{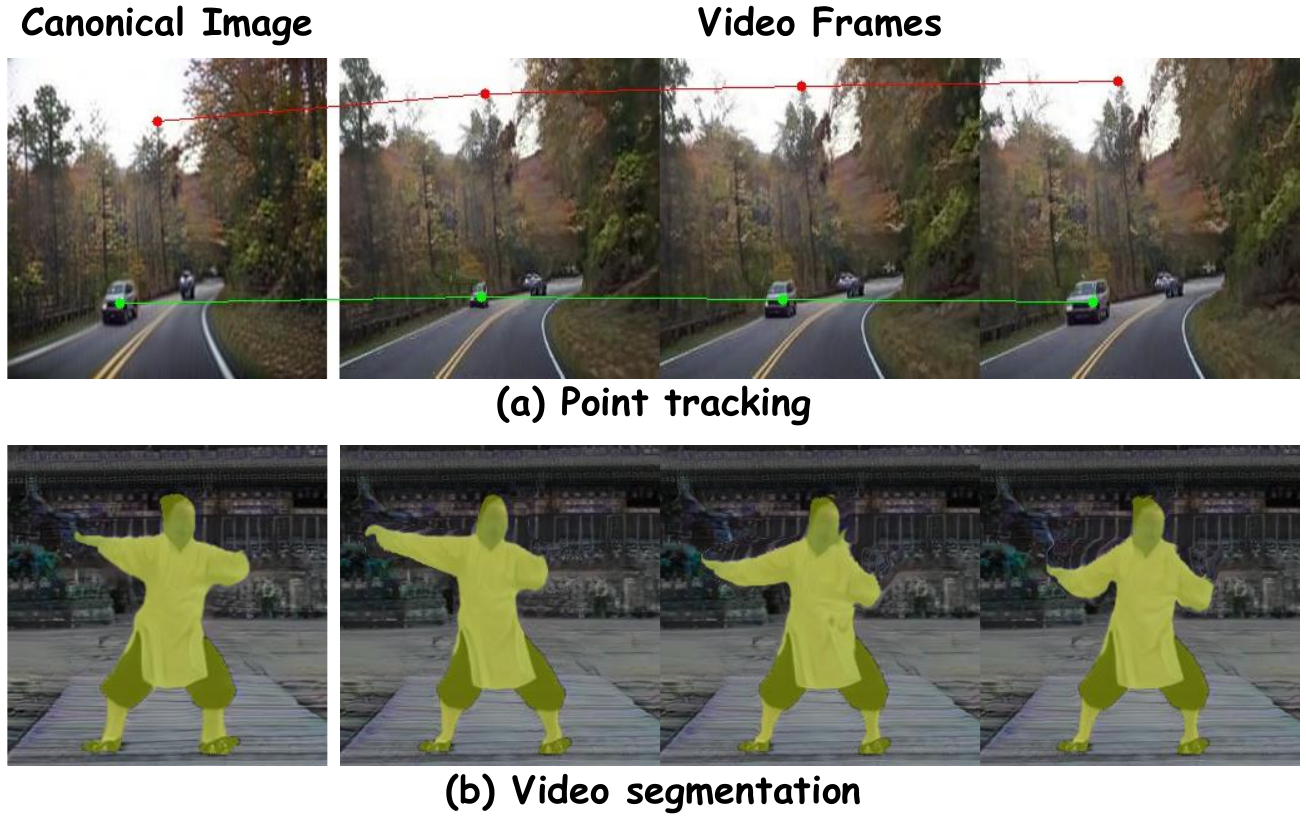}
    \captionsetup{type=figure}
    \caption{%
        \textbf{Applications on (a) point tracking and (b) video segmentation.}
        The selected points or generated masks in the canonical image can be propagated smoothly to the individual frames.
    }
    \label{fig:tracking}
\end{figure}

\noindent\textbf{Consistent video editing.}
As shown in Fig.~\ref{fig:edit}, the canonical image can serve as an interactive interface for editing. 
We first consider propagating the scribble added by users on the canonical image to the whole video. As shown in Fig.~\ref{fig:edit}(a), the scribble added on the side of the car is coherently propagated to the entire video and changes reasonably following the viewpoint. 

Second, we use ControlNet~\cite{zhang2023adding} to edit the global style of the canonical image, as shown in Fig.~\ref{fig:edit}(b), the daytime scene is edited to ``starry night" style. It can be seen that the edited video frames are temporally consistent and maintain the structure of the original video. 
Similarly, we can also use ControlNet to edit the foreground of the image, \textit{e.g.} editing a ``person" to a ``robot", as shown in Fig.~\ref{fig:edit}(c).

\noindent\textbf{Point tracking.}
The user can select points of interest on the canonical image and automatically obtain the trajectories of these points over the generated video. In implementation, we achieve point tracking by locating the pixels in the video frames that are sampled from these points of interest in the canonical image. As shown in Fig.~\ref{fig:tracking}a, our method can accurately track the left headlight of the vehicle, as well as the top of the tree of interest.

\noindent\textbf{Video segmentation.}
We can also use the deformation field to propagate segmentation masks on the canonical images to realize video segmentation. As shown in Fig.~\ref{fig:tracking}b, we use SAM~\cite{kirillov2023segment} to segment the person of interest on the canonical image, and our \method automatically outputs the high-quality video segmentation result of the person.

\noindent\textbf{Video generation of the same visual content performing different motions.}
We can generate varying deformation fields for the same visual content, as shown in \cref{fig:multi_motion}. The visual content is represented by the canonical image in (a), while two video samples with corresponding deformation fields are shown in (b) and (c), respectively. 
Take the second row as an example, the same person plays Tai Chi with his body moving in different directions, as reflected by both the deformation fields and the video frames.
\section{Conclusion}
\label{sec:conclusion}

We have proposed a novel video generation paradigm that explicitly decomposes a video into a shared canonical image and a frame-wise deformation field on a structured high-dimensional space.

Our experiments have shown 
promising results. Our method outperforms previous methods both qualitatively and quantitatively.
Furthermore, the explicit decomposition makes possible a wide range of downstream applications, including video editing, point tracking, and video segmentation.
In future work, we hope to apply the same paradigm to other representative generative models, such as the diffusion models, and scale up the training data, model size, and computation for open-world video generation.
\section*{Acknowledgements}
This work was in part supported by the National K\-e\-y R\-\&\-D Program of China (No.\  20\-22\-ZD\-01\-18\-7\-0\-0).
{
\small
\bibliographystyle{ieeenat_fullname}
\bibliography{ref.bib}
}

\appendix
\renewcommand\thefigure{A\arabic{figure}}
\renewcommand\thetable{A\arabic{table}}
\renewcommand\theequation{A\arabic{equation}}
\setcounter{equation}{0}
\setcounter{table}{0}
\setcounter{figure}{0}

\section*{Appendix}

\section{More Implementation Details}

\begin{figure*}
    \centering
    \includegraphics[width=0.88\textwidth]{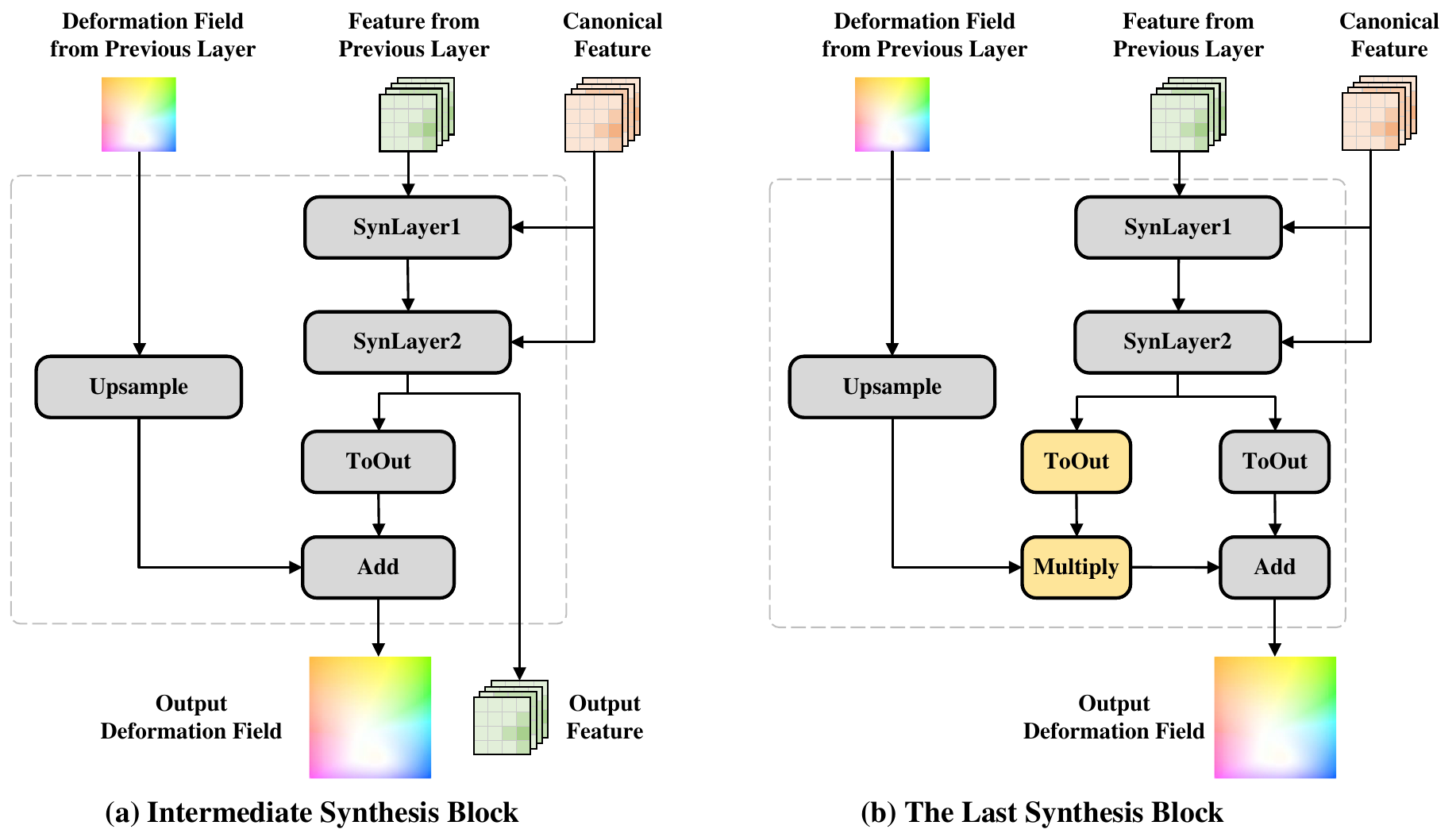}
    \captionsetup{type=figure}
    \caption{%
        \textbf{Illustration of the synthesis blocks in the deformation field generator}. (a) The intermediate synthesis block follows the structure of StylegAN2~\cite{karras2020stylegan2}, while we introduce the canonical feature to the synthesis layers (denoted as ``syn\_layer"). (b) We introduce a multiplier on the deformation field from previous layers, to realize zero initialization on the output deformation field.
        Note both synthesis layers and the output layer (denoted as ``ToOut") take the style code in $\mathcal{W}$ space for modulated convolutions~\cite{karras2020stylegan2}, which we omit for clarity.
    }
    \label{fig:supp_syn_block}
\end{figure*}

Our deformation field generator follows the structure of StyleGAN-V~\cite{skorokhodov2022stylegan}. Since it outputs the deformation field rather than an RGB image, we set the output channel to 2.
The intermediate synthesis block in our deformation field generator is shown in Fig.~\ref{fig:supp_syn_block}a. It takes as input the predicted deformation field from the previous layer, the features from the previous layer (namely the deformation feature), and the canonical feature from the canonical image generator, and outputs the updated deformation field and deformation feature. Note that in the figure we ignore the input style code from $\mathcal{W}$ space~\cite{karras2020stylegan2} for better clarity. 
As can be seen, the synthesis block contains two branches, the left branch up-samples the deformation field from the previous layer, while the right branch takes deformation features from the previous layer as inputs and passes them through two synthesis layers~\cite{karras2020stylegan2} and an output layer to obtain the residual of the deformation field, which is then summed with the up-sampled deformation field to obtain the updated deformation field. 
The ``ToOut" layer projects the deformation feature to the residual of deformation fields, which follows the structure of the ``ToRGB" layer in StyleGAN2~\cite{karras2020stylegan2}, except that we change the output channel from 3 to 2. More specifically, it consisted of a modulated convolution and an activation layer.
For the first synthesis block, there is no deformation field from the previous layer, so the results predicted by the ``ToOut" layer are directly used as the output deformation field.

\lstset{
  backgroundcolor=\color{white},
  basicstyle=\fontsize{7.5pt}{8.5pt}\fontfamily{lmtt}\selectfont,
  columns=fullflexible,
  breaklines=true,
  captionpos=b,
  commentstyle=\fontsize{8pt}{9pt}\color{codegray},
  keywordstyle=\fontsize{8pt}{9pt}\color{codegreen},
  stringstyle=\fontsize{8pt}{9pt}\color{codeblue},
  frame=tb,
  otherkeywords = {self},
}
\begin{figure}[ht]
\begin{lstlisting}[language=python]
def forward(deformation_feature, deformation_field, canonical_feature):
    # deformation_feature: (B*T, C1, H/2, W/2), where B is the batch size and T is the number of timesteps
    # deformation_field: (B*T, 2, H/2, W/2)
    # canonical_feature: (B, D, H/2, W/2), interpolated to the same resolution as the deformation_feature


    # the first synthesis layer (upsamples the feature)
    deformation_feature = syn_layer1(deformation_feature, canonical_feature)  # (B*T, C2, H, W)
    # the second synthesis layer
    deformation_feature = syn_layer2(deformation_feature, canonical_feature)  # (B*T, C2, H, W)

    # upsample deformation_field from the previous layer
    deformation_field = Upsample(deformation_field)  # (B*T, 2, H, W)

    # obtain output deformation field
    prediction = ToOut(deformation_feature)  # (B*T, 4, H, W)
    multiplier, adder = prediction.split(2, dim=1)  # both in (B*T, 2, H, W)
    deformation_field = deformation_field * multiplier + adder  # (B*T, 2, H, W)

    return deformation_field
    
\end{lstlisting}
\caption{
\textbf{Pseudo-code for the forward function of the last synthesis block in the deformation field generator.} 
Note both synthesis layers (denoted as ``syn\_layer") and the output layer (denoted as ``ToOut") take the style code in $\mathcal{W}$ space for modulated convolutions~\cite{karras2019stylegan}, which we omit for clarity.
\texttt{Upsample} denotes the up-sampling process. 
}
\label{fig:supp_code}
\end{figure}

As described in the main text, we initialize the deformation field to zeros. To this end, we introduce a new multiplier in the last synthesis block of the deformation field generator, as shown in \cref{fig:supp_syn_block}b. The output of our ``ToOut" layer is divided into a multiplication term and an addition term operating on the previous layer's deformation field, to obtain the final output deformation field. 
The detailed forward process of the last synthesis block is also illustrated in the pseudo-code in \cref{fig:supp_code}. 
In this way, we can ensure that the deformation field of the final output deformation field is initialized to zeros by simply initializing the convolution weight and bias in ``ToOut" to zeros.

Additionally, to realize our structural temporal smoothness constraint, we arbitrarily sample a time $t$ in each training iteration and subsequently take two adjacent times $t-1$ and $t+1$ to obtain the deformation fields of the three adjacent video frames and use them to compute the structural temporal smoothness constraint loss.

\section{More Analysis}

\paragraph{Ablations on deformation field initialization.}

\begin{table}[t]
\footnotesize
\caption{
    \textbf{Ablations on deformation field initialization strategy.} 
    ``\textit{w/o multi.}" indicates the default initialization following StyleGAN-V~\cite{skorokhodov2022stylegan}, without our inducted multiplier in \cref{fig:supp_syn_block}b.
    ``xavier" and ``zero" indicate applying xavier or zero initialization on the weight of the modulated convolution layer in the last ``ToOut" layer, respectively.
}
\label{table:suppp_ablate_deform_init}
\centering
\begin{tabular}{@{}lllllll@{}}
\toprule
\multirow{2}{*}{\textbf{Config}}              & \multicolumn{3}{c}{YouTube Driving} & \multicolumn{3}{c}{SkyTimelapse} \\ \cmidrule(l){2-7} 
            & FVD$_{16}$    & FVD$_{128}$   & FID           & FVD$_{16}$    & FVD$_{128}$   &  FID       \\ 
\midrule
\textit{w/o} multi.  & 282.0 & 234.4 & 9.8 & 51.7 & 163.5 & 44.7 \\ 
\midrule
xavier    & \textbf{181.3} & 203.2 & 8.5 & 42.5 & 133.6 & 41.2 \\
zero     & {181.9} & \textbf{188.5} & \textbf{8.0}       & \textbf{41.8}        & \textbf{128.1}       & \textbf{40.7} \\
\bottomrule
\end{tabular}
\end{table}

\begin{figure*}
    \centering
    \includegraphics[width=\textwidth]{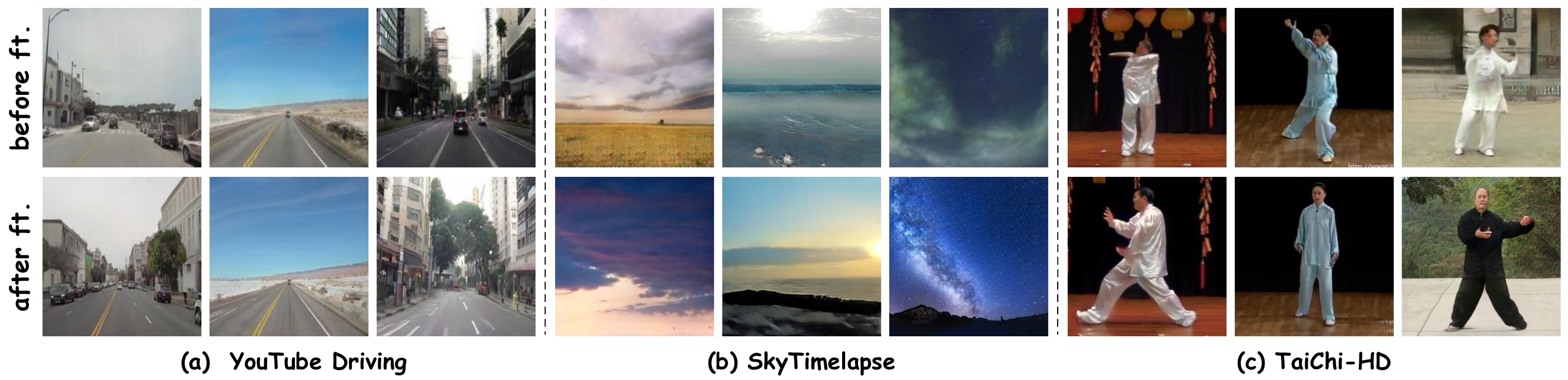}
    \caption{%
        \textbf{Image samples}. The first row presents the images generated by the pre-trained image generator, denoted as ``before ft.", while the second row shows the individual frames generated by the fine-tuned video generator, denoted as ``after ft.".
    }
    \label{fig:supp_fid}
\end{figure*}

\begin{figure}
    \centering
    \includegraphics[width=0.48\textwidth]{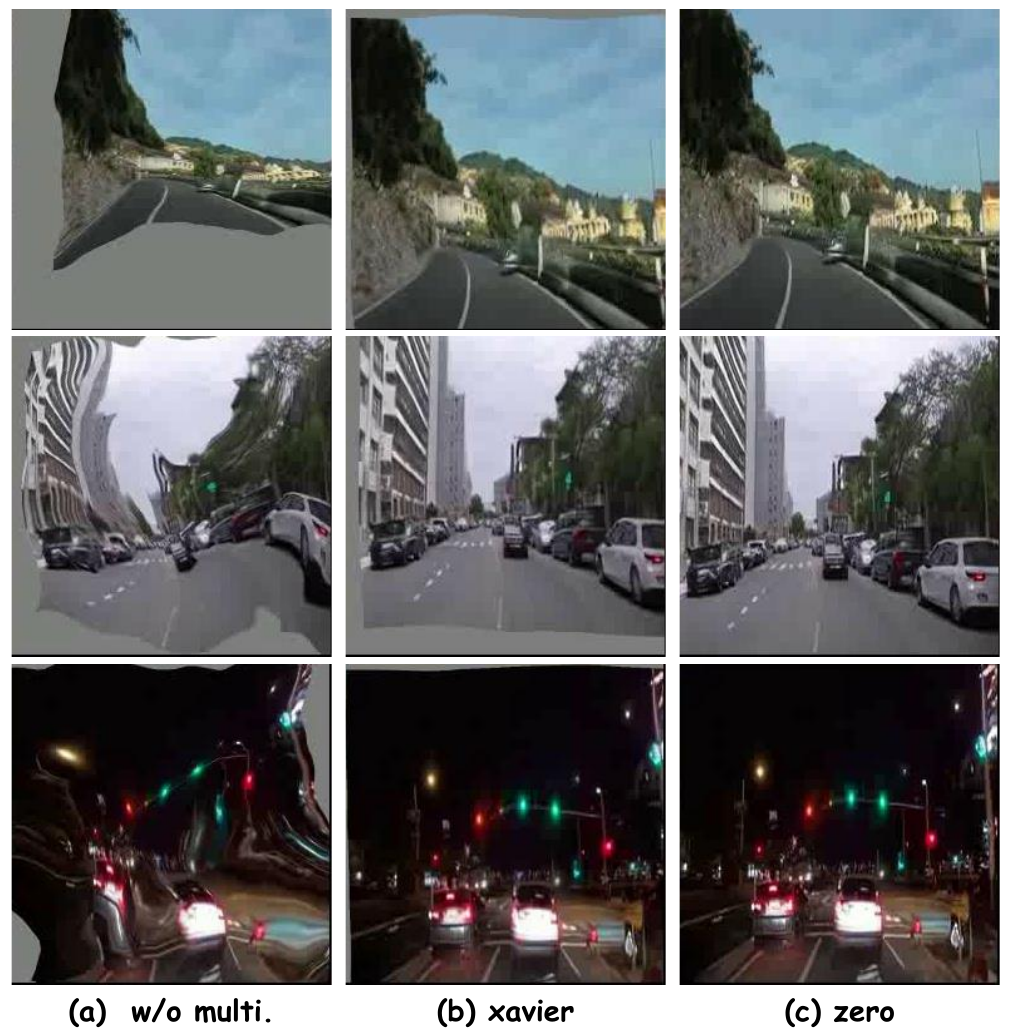}
    \caption{%
        \textbf{Visualization of different deformation field initialization strategies.} (a) Initialization without our introduced multiplier, denoted as ``w/o multi.". (b) and (c) are initialized with the multiplier, and apply xavier and zero initialization on the last ``ToOut" layer, respectively.
        Note with zero initialization, the initial video frames simply copy the canonical image generated by the pre-trained image generator.
    }
    \label{fig:supp_init}
\end{figure}

As described in the main text, we initialize the deformation field to zeros in the fine-tuning phase. In this way, the video frames are all initialized as canonical images. Here we perform ablated experiments on the deformation field initialization.  
We consider three different initialization as the following: 
(a) Using the default initialization of StyleGAN-V~\cite{skorokhodov2022stylegan} with the model structure in \cref{fig:supp_syn_block}a, without our newly introduced multiplier, denoted as ``w/o multi."
(b) using the structure in \cref{fig:supp_syn_block}b as the last layer of the synthesis block, \textit{i.e.}, introducing the multiplier term when the updating deformation field. We consider both zero and xavier initialization~\cite{glorot2010understanding} for the modulated convolution in the ``ToOut" layer, denoted as ``zero" and ``xavier", respectively. The initial video frames of different initialization methods are shown in Fig.~\ref{fig:supp_init}.

The experimental results are shown in \Cref{table:suppp_ablate_deform_init}. It can be seen that ``w/o multi." performs poorly, which is mainly due to the fact that the deformation field accumulated random initialized bias over individual synthesis blocks and is thus initialized to large random values.
As a result, the initialized video frames significantly differ from a natural image, as shown in Fig.~\ref{fig:supp_init}a, which is detrimental to model training. 
By contrast, introducing our multiplication term brings much better performance, since the initial offset of the deformation field is small. What's more, using zero initialization works slightly better.

\paragraph{Image quality before and after video fine-tuning.}

\begin{table}[t]
\footnotesize
\caption{
    \textbf{FID before and after fine-tuning on the video dataset.} ``before ft." indicates the image generative model pre-trained on video frames, and ``after ft." denotes the video generative model fine-tuned on the video dataset.
}
\label{table:supp_fid}
\vspace{2pt}
\centering
\begin{tabular}{@{}lccc@{}}
\toprule
\multirow{2}{*}{\textbf{Config}}    & \multicolumn{1}{c}{YouTube Driving} & \multicolumn{1}{c}{SkyTimelapse} & \multicolumn{1}{c}{TaiChi-HD} \\ \cmidrule(l){2-4} 
          & FID  &  FID  &  FID    \\ 
\midrule
before ft.  & 11.4 & 43.6 & 24.4 \\ 
after ft.  & \textbf{8.0} & \textbf{40.7} & \textbf{20.7} \\ 
\bottomrule
\end{tabular}
\end{table}

As described in the main text, our approach is able to reuse the pixels in the canonical image and alleviate the difficulty in producing a video. Here, we compare the quality of images generated by the pre-trained image generation model and the fine-tuned video generation model, denoted as ``before ft." and ``after ft.", respectively.
The results are shown in \Cref{table:supp_fid} and \cref{fig:supp_fid}. 
Interestingly, after fine-tuning, the video generation model produces images with better quality than the pre-trained image generation model, \textit{e.g.}, the FID is improved from 11.4 to 8.0 on the YouTube Driving dataset. we also observe better image quality from the second row in \cref{fig:supp_fid}.
We speculate that this is mainly due to the fact that our \method produces video frames by sampling pixels from the shared canonical image, instead of generating them from scratch. 
Moreover, the canonical image is shared for all video frames and can learn a better pixel representation from the whole video, compared to learning the pixel representation from a single frame image.

\section{More Qualitative Results}

\begin{figure*}
    \centering
    \includegraphics[width=\textwidth]{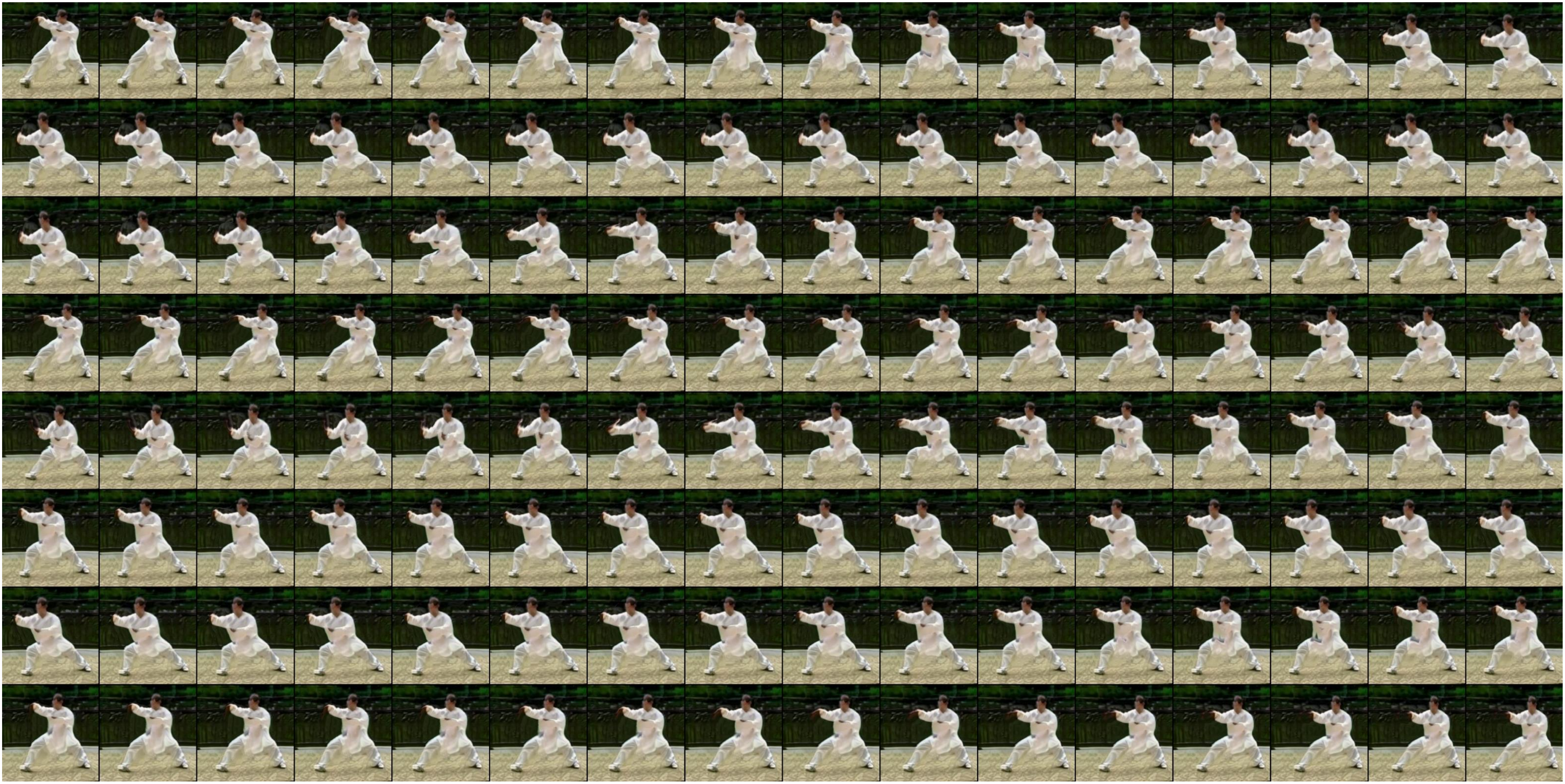}
    \captionsetup{type=figure}
    \caption{%
        \textbf{128-frame video resolution generated by \method}. Please view each row from left to right, and view the rows from the top to bottom.
    }
    \label{fig:supp_128_frame}
\end{figure*}

In \cref{fig:supp_128_frame}, we present a sampled video of full 128-frames. As can be seen, our \method produces temporally consistent video even for long video of 128 frames.

\section{Limitations and Discussions}

In this section, we discuss the potential limitations of our approach and possible solutions.

First, in our approach, both the canonical image and the deformation field are represented by a fixed-resolution grid, which is inconsistent with the property that videos are spatially continuous and may limit the model to be trained on data of arbitrary resolution. In contrast, many work has attempted to introduce implicit representations in videos~\cite{yu2022dign,codef}. Similarly, we can represent the canonical and deformation fields with a set of MLP parameters and sample points to get the video at the resolution we need.
This further improves the capacity of the canonical and deformation field representation, and enables new applications such as video super-resolution in a natural way. 

Second, our downstream applications, including video editing, point tracking, video segmentation, \textit{etc.}, rely on the canonical image and the deformation field that we obtain when generating the video.
However, in real videos, the canonical image and the deformation field cannot be directly obtained. As a result, the direct application of our method to real videos is not currently feasible. To cope with this problem, we consider two possible solutions. The first one is to inverse the real video into the canonical image and deformation field, leveraging sophisticated methods in  inversion~\cite{xia2022gan}. The second one is to train our model on the single input real video, so that the model can generate the input real video, and also naturally obtain the canonical image and deformation field during the generation process. This solution is related to video decomposition methods such as Layered Neural Atlas~\cite{lu2020layered} and CoDeF~\cite{codef}, while our pre-trained video generation models can serve as a better prior for video decomposition.

Third, our approach may be degraded when generating extremely long videos, due to the limited amount of content information that can be encoded in a single canonical image. To tackle this problem, we can consider the following solutions: 
(1) As mentioned before, we can introduce implicit representation on the canonical image to encode the content information using powerful neural networks, alleviating the problem of limited information encoded by a fixed-size image.
(2) We can introduce a refinement network to fill in the information that is absent in the canonical image after warping the canonical image with the deformation field. Similar ideas can be found in the work on image-to-video generation~\cite{ni2023conditional}.
(3) When generating long videos, we can use the already generated video frames to update the canonical image and thus introduce new video content before generating subsequent video frames.
We will explore these improvement directions in future work.

\end{document}